\documentclass[default,iicol]{sn-jnl}% Default with double column layout

\jyear{2022}%

\theoremstyle{thmstyleone}%
\theoremstyle{thmstyletwo}%
\usepackage{amsfonts}

\theoremstyle{thmstylethree}%

\begin{document}

\title[TCC]{Transformer-based Context Condensation for Boosting Feature Pyramids in Object Detection}

%%=============================================================%%
%% Prefix	-> \pfx{Dr}
%% GivenName	-> \fnm{Joergen W.}
%% Particle	-> \spfx{van der} -> surname prefix
%% FamilyName	-> \sur{Ploeg}
%% Suffix	-> \sfx{IV}
%% NatureName	-> \tanm{Poet Laureate} -> Title after name
%% Degrees	-> \dgr{MSc, PhD}
%% \author*[1,2]{\pfx{Dr} \fnm{Joergen W.} \spfx{van der} \sur{Ploeg} \sfx{IV} \tanm{Poet Laureate} 
%%                 \dgr{MSc, PhD}}\email{iauthor@gmail.com}
%%=============================================================%%

\author[1]{\fnm{Zhe} \sur{Chen}}\email{zhe.chen1@sydney.edu.au}

\author[1]{\fnm{Jing} \sur{Zhang}}\email{jing.zhang1@sydney.edu.au}

\author[1]{\fnm{Yufei} \sur{Xu}}\email{yuxu7116@uni.sydney.edu.au}

\author*[2,1]{\fnm{Dacheng} \sur{Tao}}\email{dacheng.tao@gmail.com}

\affil*[1]{\orgdiv{School of Computer Science, Faculty of Engineering}, \orgname{The University of Sydney}, \orgaddress{ \city{Darlington}, \state{NSW} \postcode{2008}, \country{Australia}}}

\affil[2]{\orgname{JD Explore Academy}, \orgaddress{\country{China}}}

%\affil[3]{\orgdiv{Department}, \orgname{Organization}, \orgaddress{\street{Street}, \city{City}, \postcode{610101}, \state{State}, \country{Country}}}

%%==================================%%
%% sample for unstructured abstract %%
%%==================================%%

\abstract{Current object detectors typically have a feature pyramid (FP) module for multi-level feature fusion (MFF) which aims to mitigate the gap between features from different levels and form a comprehensive object representation to achieve better detection performance. However, they usually require heavy cross-level connections or iterative refinement to obtain better MFF result, making them complicated in structure and inefficient in computation. To address these issues, we propose a novel and efficient context modeling mechanism that can help existing FPs deliver better MFF results while reducing the computational costs effectively. In particular, we introduce a novel insight that comprehensive contexts can be decomposed and condensed into two types of representations for higher efficiency. The two representations include a locally concentrated representation and a globally summarized representation, where the former focuses on extracting context cues from nearby areas while the latter extracts key representations of the whole image scene as global context cues. By collecting the condensed contexts, we employ a Transformer decoder to investigate the relations between them and each local feature from the FP and then refine the MFF results accordingly. As a result, we obtain a simple and light-weight Transformer-based Context Condensation (TCC) module, which can boost various FPs and lower their computational costs simultaneously. Extensive experimental results on the challenging MS COCO dataset show that TCC is compatible to four representative FPs and consistently improves their detection accuracy by up to 7.8 \% in terms of average precision and reduce their complexities by up to around 20\% in terms of GFLOPs, helping them achieve state-of-the-art performance more efficiently. Code will be released.}

%%\pacs[JEL Classification]{D8, H51}

%%\pacs[MSC Classification]{35A01, 65L10, 65L12, 65L20, 65L70}

\maketitle

\section{Introduction}
Object detection aims to detect and localize objects with various sizes on images. To detect objects of diversified sizes, researchers have found that a feature pyramid (FP), which collects features from different levels of the backbone network and performs detection in a pyramidal hierarchy, are important for high-quality detection. Using a deep convolutional neural network (DCNN) like ResNet\cite{he2016deep} which provides multiple levels of DCNN features as the backbone network, the feature pyramid, such as the FPN \cite{lin2017feature}, detect smaller objects at shallower levels and larger objects at deeper levels.  
The shallower-level features have higher-resolutions and thus are better at describing objects with smaller sizes, but these features are worse at extracting abstract and robust representation. On the other hand, the deeper-level features have much smaller scales and are better for objects with larger sizes, while they lose too many local details for detection. Although feature pyramids can achieve more accurate detection by enabling the detection of smaller objects on shallower-level features and larger objects on deeper-level features, using either lower-level or higher-level features alone still cannot satisfy high-quality detection demand. In the related research, it shows that cooperating the advantages of features from different levels by performing multi-level feature fusion (MFF) is very helpful to obtain more descent representation of objects and achieve better detection performance.

\begin{figure}[t]
%\centering
\includegraphics[width=\linewidth, height=0.28\textheight]{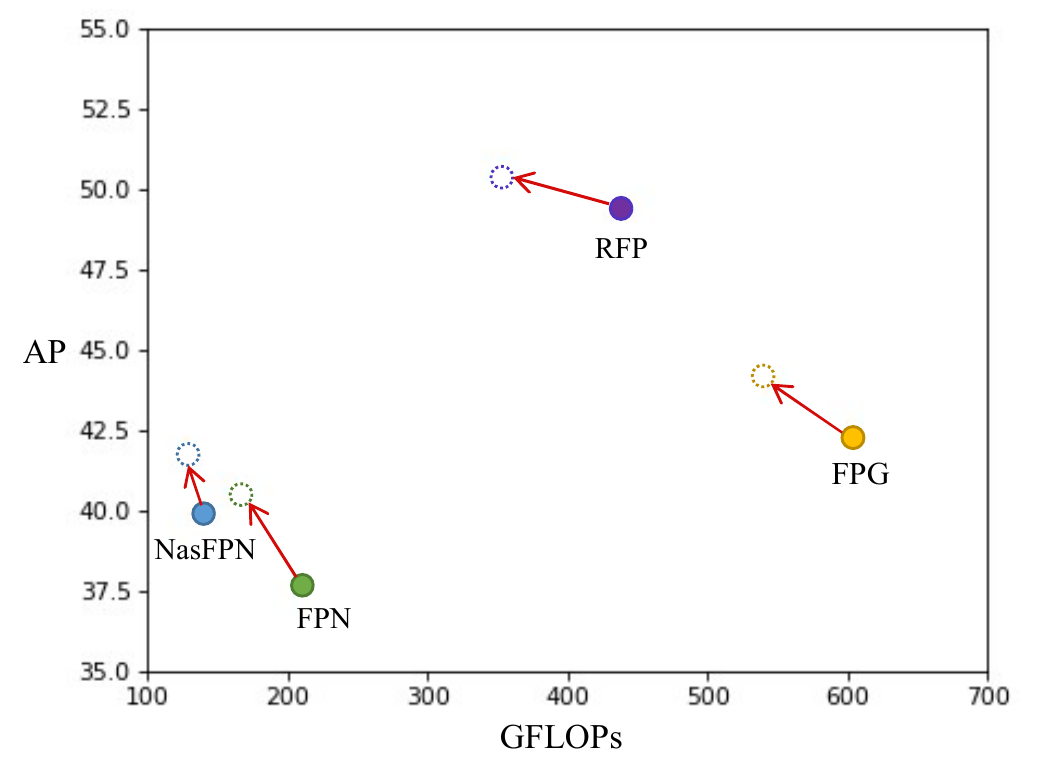}
  \caption{The proposed Transformer-based Context Condenser (TCC) (dotted circles) can effectively improve the multi-level feature fusion in different baseline feature pyramids (solid circles) and the final detection performance by utilising rich context information. In the meantime, the TCC also reduces the required computational complexity promisingly by condensing the contexts into two simplified representations, including a locally concentrated representation and a globally summarized representation. }%
  \label{fig:title}
\end{figure}

To achieve effective MFF, that gaps between features of different levels are supposed to be mitigated for better fusion. However, we found that current popular mechanisms usually require excessive computational complexity to ensure more promising MFF results. For example, in the typical feature pyramid as introduced in FPN \cite{lin2017feature}, the MFF is fulfilled by a normal 3 by 3 kernel based convolution operation after adding each shallower-level feature with the upsampled adjacent deeper-level feature. This design is simple but very primitive. It can be quite difficult for only using a 3 by 3 convolution to effectively bridge the gap between different feature representations to refine the final detection results promisingly. Nevertheless, the 3 by 3 convolution itself requires plenty of extra complexity, especially for shallower levels with high resolutions.
Derived from FPN, researchers further developed more complicated and also more costly architectures to help achieve better MFF. For example, the authors of NasFPN\cite{ghiasi2019fpn} applied an auto-network searching technology \cite{zoph2016neural,zoph2018learning,baker2016designing} to make the machine automatically search for a better feature pyramid structure and better MFF strategy. The searched feature pyramid is then iterated for multiple times in a successive order to further refine the MFF results, which still brings considerable computational complexity. Later, the FPG \cite{chen2020feature} and RFP \cite{qiao2021detectors} explored even more comprehensive and complicated MFF strategies in pyramid architectures. Although these methods achieved state-of-the-art performance, they continues to introduce multiple iterations of the same processing structure to ensure more effective MFF in feature pyramids, making them still very computational costly. 
Instead of repeated and exhaustive feature modeling, we find that contexts could be of great benefit for refining the fused representation after MFF. More specifically, by considering extra visual cues from a larger surrounding area as contexts, it could become easier for shallower-level features to resist visual noises and for upsampled deeper-level features to improve local detailed descriptions. As a result, the gap between features from different levels could be alleviated effectively with contexts. Some studies \cite{zhao2017pyramid,chen2021recursive} have demonstrated the effectiveness of including contexts for improving detection and other visual understanding tasks. However, a more comprehensive context modeling is usually still accompanied with a higher computational complexity, which would limit its advantages for boosting feature pyramids efficiently. 

Although comprehensive context modeling may be still costly, we further find that the rich contexts can be decomposed and simplified to avoid the need of heavy computation. We are motivated by the studies \cite{uzkent2020learning,zhang2014part,chen2019drop} that successfully summarized complicated visual appearances using a concrete representation and some key representations. In those studies, the concrete representation is used to represent major visual patterns in images, and the key representations summarize overall visual patterns into a few key features. 
By making key representations tend to complement concrete representation based on the relations between both types of representations, jointly using both types of representations can avoid exhaustive feature processing without sacrificing much visual understanding performance. Accordingly, we propose that the rich contexts can also be decomposed into a concrete representation and some key representations to achieve efficient modeling. We tend to term such decomposition of contexts as context condensation.

To condense and utilize contexts for boosting MFF and feature pyramids efficiently, in this study, we propose a novel Transformer-based context condensation (TCC) module for object detection. After adding features from different levels as fusion results, the TCC first condenses rich contexts on the fused representations and then use the condensed contexts to refine the feature of each entry on the fused feature map with the help of a Transformer decoder. More specifically, in a TCC module, we attempt to first condense contexts by decomposing comprehensive context information into a locally concentrated representation and a globally summarized representation. We make the locally concentrated representation provide concrete representation of local contexts, and we attempt to summarize global contexts into a few key features whose locations are predicted by an extra small neural network. Then, we collect the feature of locally concentrated representation and the features of globally summarized contexts as the condensed context representation. With the condensed contexts, we apply a Transformer decoder to estimate the relations between the condensed contexts and each feature to be refined, so that more related and useful context information can be identified and highlighted from condensed contexts. Based on the estimated relations, the Transformer decoder translates the obtained condensed context features into a more appropriate contextual representation to achieve better MFF refinement. Therefore, using the condensed contexts, the exhaustive modeling of comprehensive contexts on the feature map can be reduced to the modeling of only a few context features. The TCC can thus achieve promising MFF refinement performance without introducing too large complexities. Fig. \ref{fig:title} shows the performance improvement and complexity reduction of applying our proposed TCC on MFF of different typical feature pyramids.

The major contributions of this work can be summarized as follows:
\begin{itemize}
\item A novel Transformer-based Context Condensation (TCC) module is presented to boost MFF in feature pyramids for detecting object effectively and efficiently. 
It mainly helps MFF achieve better detection performance without requiring great computational complexities. 
\item The proposed TCC module is easy-to-inject and can boost MFF in many different popular feature pyramid methods developed for object detection. It is also compatible with different backbone networks and is not sensitive to specific DCNN architectures. It can be trained with different feature pyramids in an end-to-end manner. 
\item Comprehensive empirical studies on MS COCO detection dataset \cite{lin2014microsoft} show that our proposed TCC module can reduce the typical FPN method by more than 30 GFlops while achieving around 7.8 \% relative improvement. By applying the TCC module to state-of-the-art feature pyramids like NasFPN\cite{ghiasi2019fpn} and RFP\cite{qiao2021detectors}, our method consistently reduces complexity and improves detection performance, accessing compelling performance on the COCO dataset. Codes shall be released upon acceptance of this paper. 
\end{itemize}

\section{Related Work}

\subsection{Feature Pyramids for Object Detection}
In computer vision, building pyramidal representations which describe visual appearances at different levels or resolutions have shown to be effective for tackling scale variation problem\cite{chen2020feature}.  
In modern deep learning-based object detectors, feature pyramids have become a core component. Early works like SSD \cite{liu2016ssd} and MS-CNN \cite{cai2016unified} extract and utilize multi-level feature maps to perform detection, but they do not consider feature fusion or aggregation across different levels, which makes them sub-optimal for detection. Then, the most popular feature pyramid for detection is introduced in the FPN\cite{lin2017feature}. It presents a simple multi-level feature extraction and fusion architecture, which consists of several top-down pathways and lateral connections, to help distribute the detection of objects with different sizes on features from different levels of the backbone DCNN. Similar to the design logic of FPN, many methods \cite{kong2017ron,huang2017multi,liu2018path, sun2019deep,yu2018deep, zhao2019m2det} have been proposed to achieve more effective pyramidal feature extraction by developing more comprehensive and more complicated multi-level feature extraction and fusion structures. In addition to these hand-crafted feature pyramids, the work of NasFPN\cite{ghiasi2019fpn} follows the NAS algorithms \cite{zoph2016neural,zoph2018learning,baker2016designing} which enables automatic searching for more efficient and more effective network architectures to facilitate the automatic searching for a better feature pyramid for object detection. The NasFPN achieves promising improvements over the FPN, but it requires several stacks of the same feature pyramid structure to refine the multi-level features. Later, Auto-FPN\cite{xu2019auto} has introduced a more comprehensive searching policy to improve detection, but its searched architecture has different structures in different stacks, making it less scalable. More recently, researchers have developed methods like FPG\cite{chen2020feature} and RFP\cite{qiao2021detectors} to achieve more comprehensive feature fusion in different levels of several stacks of feature pyramids. Although state-of-the-art performance can be accessed by these methods, they still introduce much extra computation complexity for object detection.

\subsection{Context Modeling for Object Detection}
Visual contexts have been demonstrated to be important for visual perception. Researchers have introduced additional visual cues as the useful contextual information to facilitate the description of visual appearance at each particular location. In object detection, contexts have been used for improving the representation of convolutional features extracted at different levels of the backbone DCNN. For example, many methods \cite{gidaris2015object,komodakis2016attend} sampled extra windows with different scales around each region of interest (RoI) to extract surrounding visual features that can improve the later classification and regression operations in detectors.
In addition to context windows, some other studies \cite{zeng2017crafting, chen2017spatial} successfully improved the features for describing objects of different sizes by introducing recurrent neural networks to encode contextual information in larger areas. Authors of \cite{chen2021recursive} further proposed a comprehensive and dynamic context modeling framework to improve multi-level feature and boost object detection performance effectively. Besides, the non-local operation \cite{wang2018non} shows great benefits for improving detection by modeling all the sub-sampled features extracted from input images.
However, current cutting-edge context modeling methods would usually introduce excessive computational complexity to model contexts more comprehensively, making it difficult to apply current context modeling methods to augment multi-level feature fusion in feature pyramids for object detection. 

\begin{figure*}[t]
\centering
\includegraphics[width=\linewidth]{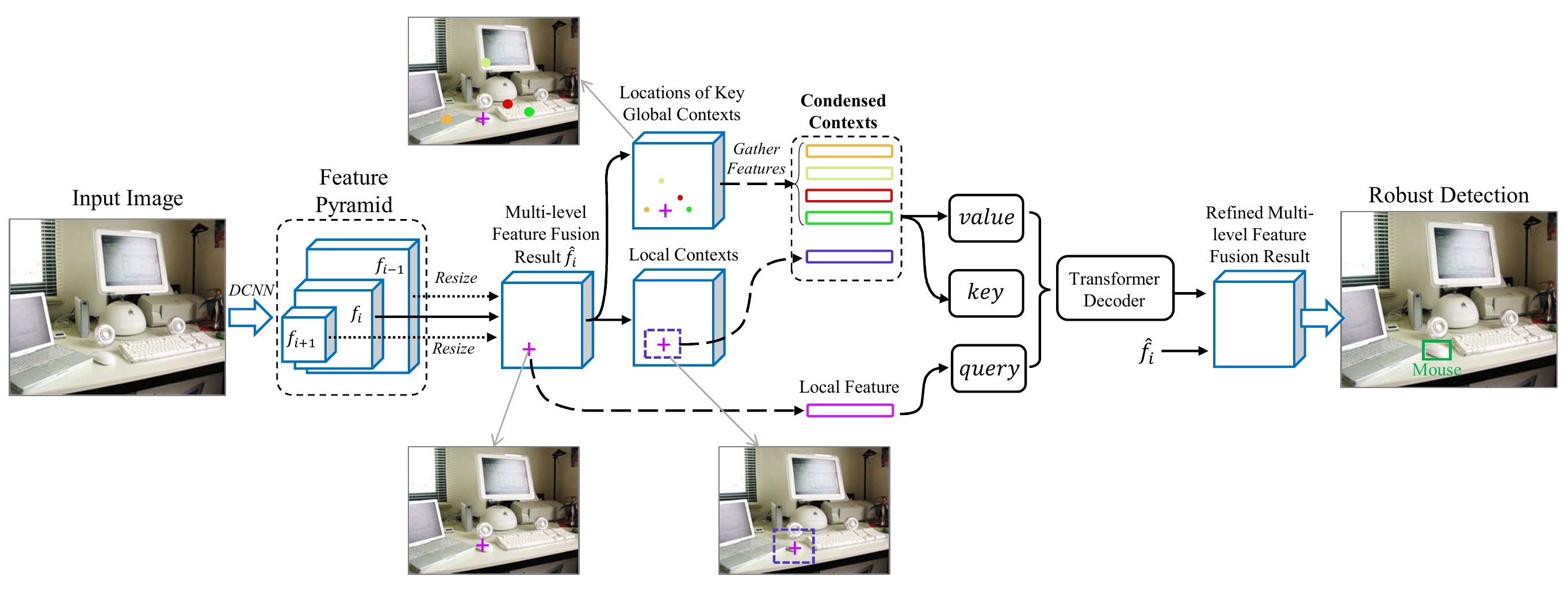}
  \caption{Processing pipeline of the proposed Transformer-based Context Condensation (TCC) module for boosting multi-level feature fusion. In the figure, the purple cross indicates a local entry on the fusion results at level $i$, the dots with different colors indicate predicted locations of key globally summarized context features, and dashed deep purple rectangular represents locally concentrated context feature. Solid arrows, dotted arrows, dashed arrows, and light grey arrows represent convolutions, resizing operations, feature gathering operations, and corresponding locations on the input image, respectively. }%
  \label{fig:main}
\end{figure*}

\subsection{Decomposition of Feature Representation}
Researchers have found that complicated visual patterns can be decomposed into different types of simplified representations.
For example, Zhang \textit{et al.} \cite{zhang2014part} introduced part-based representations to complement the holistic representation for fine-grained classification. The authors of \cite{uzkent2020learning} used part-global representation learn when and where to zoom images for recognition. The study \cite{chen2019drop} represented image features with information from low and high spatial frequency domains. However, we found in these studies that the correlation between the general and specific features lacks sufficient study, thus it could become difficult make networks decide which specific features are more beneficial for complementing general features. We argue that the correlation between different types of simplified features is important to achieve promising performance. In particular, the level of correlation or agreement between different features can indicate their relevance and correlation reveal whether involved features share the similar and complementary information that is of interests. Although researchers have developed cosine similarity-based \cite{wojke2018deep}, Euclidean distance-based \cite{qi2017pointnet}, Intersect-over-Union (IoU) based \cite{chen2018context}, and also learning-based \cite{hu2018relation,chen2017spatial} correlation estimation methods for various vision tasks, these methods either cannot be directly applied on the general and specific features or could be difficult to describe complicated relations. Instead, inspired by the Transformers for vision \cite{dosovitskiy2020image, liu2021swin, xu2021vitae} that show to be extremely powerful at modeling complicated relations between features, we propose a Transformer-based context encoder to model specific features and learn to better complement the general features.

\subsection{Transformer in Object Detection}
Recently, the Transformer has achieved great success in computer vision\cite{dosovitskiy2020vit, touvron2021training,wang2021fp}. Researchers also developed a Transformer-based object detector, named DETR \cite{carion2020end}. It effectively takes advantage of the powerful relation modeling in Transformer for high-quality detection. However, the pure Transformer-based detector generally suffers from excessively long training periods. To address this problem, many researchers found that locality modeling is important for accelerating the training of DETR \cite{tay2020efficient, xu2021vitae, zhang2022vitaev2}. For example, the Deformable DETR \cite{zhu2020deformable} applied deformable operations \cite{dai2017deformable} to better focus on a few local areas in the Transformer. The method SMCA \cite{gao2021fast} introduced multi-scale co-attention to improve DETR with refined local representations. In addition, other studies like Conditional DETR \cite{meng2021conditional} and Anchor DETR \cite{wang2021anchor} tend to improve the spatial embedding in Transformer to help accelerate training. These two methods enhance the locality modeling of Transformer by making attention focus on potentially valuable areas on the image learned with positional embeddings. The authors of \cite{chen2022recurrent} introduces recurrent refinement to improve the locality modeling of DETR. 
Despite progress, the Transformer-based methods require a long training period to achieve promising performance, while our method aligns with normal DCNN-based object detectors and requires shorter training periods to converge.

\section{Method}
As mentioned previously, we introduce Transformer contexts condensation (TCC) modules to refine MFF and boost different feature pyramids for object detection without introducing too much computational overhead. In the TCC, condensed contexts are first collected and then a Transformer decoder is employed to translate the collected contexts into a better representation for refining features. 
Fig. \ref{fig:main} shows an overview of the proposed method. 

In the following sections, we will subsequently describe in details about our formulation of MFF operation in a feature pyramid, how we collect condensed contexts to improve MFF in the feature pyramids, and how we use Transformer to decode the collected contexts. 

\subsection{MFF in a Feature Pyramid}
In general, feature pyramids used in object detectors usually have the following processing steps, \textit{i.e.} collecting multi-level features from the backbone DCNN, re-sampling features from another level to match the scale of features at the current level, and the MFF that fuses and refines the collected multi-level features to deliver descent feature representation for detection at the current level. 

Mathematically, we use the symbol $f_i$ to represent the $i$-th level feature extracted from the backbone DCNN and use the symbol $\mathcal{S}_{j\rightarrow i}[f_j]$ represents a re-sampling function that resizes the scale of the $j$-th level feature $f_j$ into the scale of $f_i$. Then, we can summarize the MFF of the $i$-th level in the feature pyramids as the following operation:
\begin{equation}
    \hat{f}_i = g_r(f_i + \sum_{j\in \mathcal{N}(i)} \mathcal{S}_{j\rightarrow i}[f_j]),
\label{eq:fpn}
\end{equation}
where $\mathcal{N}(i)$ refers to the other levels considered for fusion at level $i$, $g_r$ refers to the feature refinement function used in different methods, and $\hat{f}_i$ is the obtained MFF result for level $i$ in the feature pyramid. For example, in the typical FPN\cite{lin2017feature}, $\mathcal{N}(i)$ is defined as its adjacent deeper level, \textit{i.e.} $\mathcal{N}(i) = i + 1$. Meanwhile, the FPN employs a simple 3 by 3 convolution to implement $g_m$ for refining fused features. Some methods like FPG\cite{chen2020feature} and RFP \cite{qiao2021detectors} have introduced more complicated mechanisms to improve $g_m$, while they usually requires several repeats of the same feature processing procedure. In practice, these methods could introduce considerably large extra computational complexity for object detectors.

\subsection{TCC}
Alternative to existing MFF mechanisms in feature pyramids, we introduce a Transformer-based context condensation (TCC) module to implement $g_r$ in Eq. \ref{eq:fpn} for improving the feature representation obtained after fusion through addition. As mentioned previously, the TCC first collect condensed contexts and then decode them for better refinement. As a result, the general mathematical description about TCC re-writes the Eq. \ref{eq:fpn} as follows:
\begin{equation}
    \left\{
    \begin{array}{ll}
       \hat{f}_i = Trans(q=\tilde{f_{i}}; k/v=g_{cc}(\tilde{f}_{i})),\\
      \tilde{f}_{i} = f_i + \sum_{j\in \mathcal{N}(i)} \mathcal{S}_{j\rightarrow i}[f_j],
    \end{array}
  \right. 
  \label{eq:tcc}
\end{equation}
where $Trans$ refers to a Transformer decoder module, the $q,k,v$ represent the query, key, and value of the Transformer, and $g_{cc}$ refers to the collection of condensed contexts.

\subsubsection{{Condensed Context Collection}}
To take advantage of rich contexts without introducing excessive computational complexity, we propose to condense contexts by decomposing them into a locally concentrated representation and a globally summarized representation that consists of a few key features. As mentioned previously, we make the locally concentrated representation describe a concrete description about nearby contexts and make the globally summarized representation delivers simplified descriptions about informative cues in the whole scene. Accordingly, we can formulate the condensed context collection as:
\begin{equation}
    g_{cc}(\tilde{f}_i) = [\tilde{f}^{lr}_i, \tilde{f}^{gr}_i],
\end{equation}
where $[\cdot]$ represents a feature concatenation operation, $\tilde{f}^{lc}_i$ and $\tilde{f}^{gc}_i$ refer to the locally concentrated representation and the globally summarized representation, respectively.

To extract the locally concentrated representation as local contexts, we can simply use a dilated convolution operation. However, different from common dilated convolutions, here we only use a very small channel number and a small dilation rate to describe local appearances efficiently. Considering that the $\tilde{f}^{lr}_i$ is the concrete representation for $i$-th level, we have the following formulation:
\begin{equation}
    \tilde{f}^{lr}_i = DilatedConv(\tilde{f}_i),
    \label{eq:gr}
\end{equation}
where $DilatedConv$ means a dilated convolution operation. To avoid excessive computational costs, the used channel number is calculated by $8 \times 2^i$ and the dilation rate is empirically set as 2.

To extract the globally summarized representation, we attempt to use a few key features to summarize and represent useful information from the whole scene. We hypothesize that the objects in the same visual scene on an input image usually shares the interests on similar global contextual information. For example, when detecting person and sheep in the same image, the sky or the meadow in the same image can facilitate the recognition of both types of objects in this image. Our experiments can validate that these shared key features can improve the overall detection performance effectively. To identify the key features from the scene, we tend to make networks automatically learn to discover which key features are more useful. In particular, we mainly perform a two-step extraction procedure: we first employ a small network to predict the locations of these key features, and then we gather these features according to the predicted locations as the summarized global contexts. 
Recall that $\tilde{f}^{gr}_i$ represents the desired globally summarized representation for the $i$-th level. Given a location set $P_i = \{(x,y)_{i,1}, \ldots, (x,y)_{i,n}\}$ describing the predicted locations of key global features, we then have the summarized global contexts according to:
\begin{equation}
        \tilde{f}^{gr}_i =  \Phi(\tilde{f}_i, P_i)
    \label{eq:gc}
\end{equation}
where $\Phi(\tilde{f}_i, P_i)$ is the function that gathers features from input feature map $\tilde{f}_i$ at the locations defined in the set $P_i$. The function $\Phi$ is further defined as:
\begin{equation}
    \Phi(\tilde{f}_i, P_i) = [\phi(\tilde{f}_i, (x,y)_{i,1}), \ldots, \phi(\tilde{f}_i, (x,y)_{i,k})\}],
    \label{eq:gcc}
\end{equation}
where $\phi(\tilde{f}_i, (x,y)_{i,k})$ is to collect detailed visual feature at the location of $(x,y)_{i,k}$.

To obtain an appropriate location set $P_i$ for collecting key global context features, we make networks learn to predict the location set automatically. To achieve this, we introduce importance scores for locations on the feature map and build the location set by collecting locations with the maximum importance scores. Suppose we collect $n$ key global context features, then we make the importance scores have $n$ dimensions (or $n$ elements for each location on the feature map). Further suppose importance scores are denoted by $\mathcal{C}_i=\{c_{i,k}\mid k=1,2,\ldots,n\}$ where $c_{i,k}$ represents the $k$-th importance score map. In a TCC module, we compute $\mathcal{C}_i$ using convolutions according to:
\begin{equation}
    \mathcal{C}_i = Split(Conv(\tilde{f}_i), n),
    \label{eq:c}
\end{equation}
where $Conv$ represents a 1x1 convolution operation with a channel number of $n$, and $Split(\cdot)$ means the operation that splits the obtained feature map into $n$ different feature maps along channel dimension. Each split feature map $c_{i,k}$ is 1-dimensional and can be considered as the importance score map for indicating the location of the desired $k$-th key global context feature. 

With the calculated and split importance score maps, we collect $P_i$ whose corresponding importance scores have the highest values in its related score maps. For example, when collecting the location of $k$-th key global context feature, we perform global max-pooling operation on the feature map $c_{i,k}$ to help identify the location $(x,y)_{i,k}$ of interest and the corresponding maximum importance score $s_{i,k}$:
\begin{equation}
    s_{i,k}; (x,y)_{i,k} = MaxPool(c_{i,k}), 
\end{equation}
where $MaxPool$ referes to the global max pooling operation that returns the pooled maximum value $s_{i,k}$ and its location $(x,y)_{i,k}$. 

To properly supervise the training of importance scores $\mathcal{C}_i$, we multiply the sigmoid-squashed maximum scores with the corresponding features collected as key global features. That is, in addition to original feature, we further formulate the collection operation $\phi(\tilde{f}_i, (x,y)_{i,k})$ as:
\begin{equation}
    \phi(\tilde{f}_i, (x,y)_{i,k}) = \tilde{f}_i\mid_{(x,y)_{i,k}} \cdot \sigma(s_{i,k}),
\end{equation}
where $\tilde{f}_i\mid_{(x,y)_{i,k}}$ means the specific feature on the feature map of $\tilde{f}_i$ located at $(x,y)_{i,k}$, and $\sigma$ represents the sigmoid function.

\subsubsection{Transformer-based Condensed Context Decoder}
With the collected condensed contexts, it is then important to incorporate the contextual information into the MFF results for refinement. Since the Transformer is powerful at modeling complicated relationships between different features and fusing features according to the estimated relationships, we apply a Transformer decoder to correlate each local feature to be refined and its corresponding condensed contexts and then translate the collected condensed contexts into a better representation for refining the fusion results of MFF.  
It is worth mentioning that we do not apply the encoder in a typical Transformer structure because we found in the experiments that this module delivers marginal improvement but extra costs in our proposed TCC module.

In the typical Transformer decoder architecture \cite{vaswani2017attention, carion2020end}, a cross-attention structure is introduced for modeling complicated relationships between different visual features. The cross-attention estimate correlation between the input \textit{query} and \textit{key} and then aggregate \textit{value} features according to the estimated correlations. In the proposed TCC modeling module, we formulate each local feature as \textit{query} and then describe the collected specific representations as both \textit{key} and \text{value}. 

We follow the typical Transformer decoder but make some minor modifications to save computation costs. As described in Eq.\ref{eq:tcc}, we suppose that each local entry of the MFF result $\tilde{f}_i$ is the query $q$ with $C$ as its feature dimension. We unify the key and value and consider condensed contexts $g_{cc}(\tilde{f}_i)$ as keys and values. Each feature of condensed contexts also has $C$ feature dimensions. We stack all the condensed contexts, include locally concentrated representation $\tilde{f}^{lc}_i$ and the globally summarized representation $\tilde{f}^{gc}_i$, into a unified tensor $F^{cc}_i$: $F^{cc}_i=g_{cc}(\tilde{f}_i)$. With $\tilde{f}_i$ and $F^{cc}_i$:, we implement the Transformer decoder, abbreviated as "$\mathcal{D}$", according to:
\begin{equation}
    \mathcal{D}(\tilde{f}_i, F^{cc}_i) = W_\mathcal{A} (\tilde{f}_i + \mathcal{A}^i F^{cc}_i),
    \label{eq:ma}
\end{equation}
where $W_\mathcal{A}$ is a trainable linear projection matrix and $\mathcal{A}^i$ represents the attentional weight:
\begin{equation}
    \mathcal{A}^i = Softmax(\frac{\tilde{f}_i (F^{cc})^T}{\sqrt{C}}).
\end{equation}

\subsubsection{Discussions}
Although our introduction of the condensed contexts is motivated by the studies \cite{uzkent2020learning, chen2019drop} which show that rich visual details can be significantly simplified with the help of a concrete representation and a few key representations, we would like to clarify that our approach is significantly different from these methods. For example, existing methods do not rely on importance score-based key location prediction to summarize global contexts. Moreover, existing methods generally require additional supervision for training, while our method learns automatically and maintains the end-to-end processing structure with different feature pyramids and object detectors. Lastly, current methods do not investigate the relations between condensed contexts and the local representation for identifying potentially useful information with the help of Transformer.

\begin{figure*}[t]
\includegraphics[width=\linewidth]{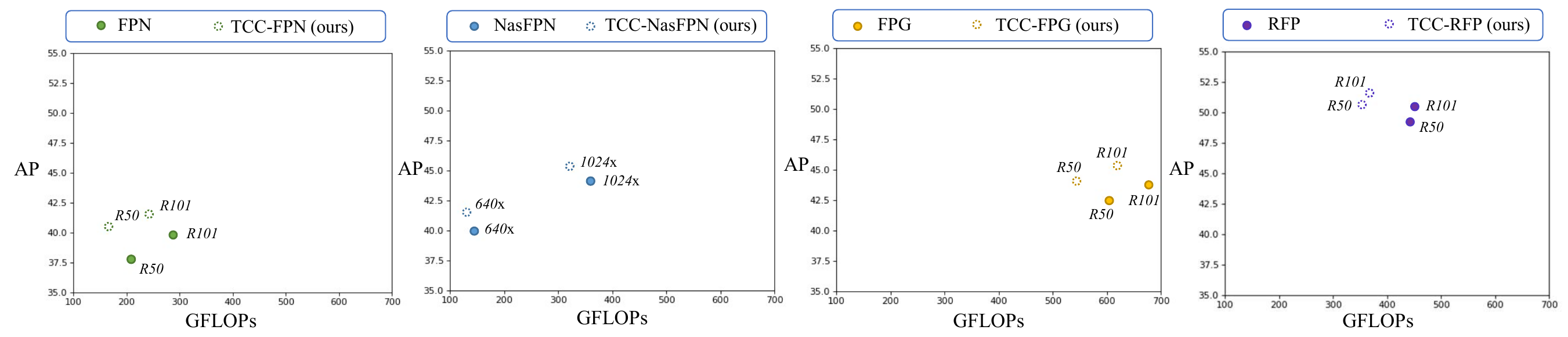}
  \caption{Accuracy in terms of AP against computational complexity in terms of GFLOPs of whether using the proposed TCC on multi-level fusion of different feature pyramids. The experiments are performed on the \textit{val} set of MS COCO \cite{lin2014microsoft} and the compared feature pyramids include FPN\cite{lin2017feature}, NasFPN\cite{ghiasi2019fpn}, FPG\cite{chen2020feature}, and RFP \cite{qiao2021detectors}. }%
  \label{fig:ap-gflops}
\end{figure*}

\begin{figure*}[t]
\includegraphics[width=\linewidth]{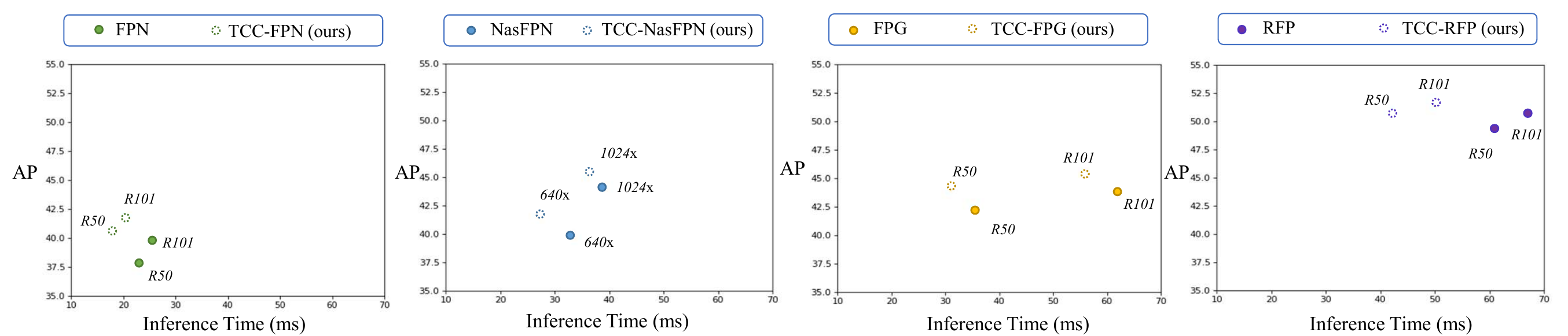}
  \caption{Accuracy in terms of AP against inference speed in terms of inference speed (ms) of whether using the proposed TCC on multi-level fusion of different feature pyramids. The experiments are performed on the \textit{val} set of MS COCO \cite{lin2014microsoft} and the compared feature pyramids include FPN\cite{lin2017feature}, NasFPN\cite{ghiasi2019fpn}, FPG\cite{chen2020feature}, and RFP \cite{qiao2021detectors}. }%
  \label{fig:speed}
\end{figure*}

\subsection{Other Implementation Details}
The TCC is easy-to-plug-in for MFF in different feature pyramids. Besides, the TCC can also be added to the features both after and before fusion. Adding TCC before fusion can bring 0.4 increase in AP point for MS COCO \textit{val} dataset comparing to only adding TCC after the fusion for MFF. In this paper, we mainly adopt the implementation of adding TCC for both before and after fusion in the experiments. 
In addition, we also apply the other implementation details as follows. Firstly, remind that we reduce the input channel dimension computed by $8\times 2^i$, which also means that we use 64 times smaller channel numbers than the input features from backbone network when using the backbone like ResNet50\cite{he2016deep}, and the level $i$ considered here ranges from 0 to 3. Secondly, we apply a residual connection between the TCC result $\hat{f}_i$ and the MFF result $\tilde{f}_i$ after Eq. \ref{eq:tcc} to perform the MFF refinement. It is also worth mentioning that we discard the multi-head structure for higher efficiency. We found that this does not sacrifice much performance. Lastly, when improving MFF before or after fusion, we stack 2 TCC modules together to achieve better refinement performance Then, in each TCC module, we predict 4 key locations to extract globally summarized context features. The ablation study of these hyper-parameters can be found at the experiment section. 

According to the implementation as described above, the computational complexity of our TCC module would not be very large by performing feature dimension reduction. In general, the complexity of the TCC is mainly from the extraction of condensed contexts and the correlation of two different condensed contexts for refinement. For condensed context extraction, the locally concentrated representation is light-weighted because of the reduced feature dimensions, and the globally summarized representation would also not consuming much computational resources because we only use a few (\textit{e.g.} 4) key points. We can show in the experiments that adding more global key points can quickly reach saturation in improvement. Regarding correlating both types of condensed context for refining MFF, the Transformer structure we used would still not bring great computational burdens due to the following designs. For example, we only employ decoders without involving encoders which we found in the experiments only have trivial impacts on the performance. Besides, since the considered feature dimension and the number of key global features are both small, it is quite efficient for obtaining the attentional weights with Transformer.

\section{Experiments}

\begin{table*}[t!]
\centering
\caption{Effects of TCC on different feature pyramids using more complicated backbone networks on the MS COCO \textit{text-dev} split, comparing to other state-of-the-art object detectors. All results of feature pyramids are reproduced using released codes. Best results are in \textbf{bold}.}
\resizebox{\linewidth}{!}{
\begin{tabular}{c| c |c| c|cc  c c c | c | c }
\hline
Feature Pyramid & Detector & Backbone & AP & AP$_{50}$ & AP$_{75}$ & AP$_{S}$ & AP$_{M}$ & AP$_{L}$ & GFLOPs & Params (M) \\ %
\hline
\hline
FPN\cite{lin2017feature} & FRCNN\cite{ren2015faster} & R101\cite{he2016deep} &36.2 &59.1 &39.0& 18.2 &39.0 &48.2&283&61\\
FPN\cite{lin2017feature} & RetinaNet \cite{lin2017focal}& R101\cite{he2016deep} &39.1& 59.1 &42.3 &21.8& 42.7 &50.2&315 &57\\
FPN\cite{lin2017feature} & FCOS\cite{tian2019fcos}&X101\cite{xie2017aggregated}&44.7& 64.1& 48.4& 27.6 &47.5& 55.6& 439 & 90\\
FPN\cite{lin2017feature} & HTC\cite{chen2019hybrid} & X101\cite{xie2017aggregated} & 47.1 &63.9 &44.7& 22.8& 43.9& 54.6 & 521 & 99\\
Hourglass \cite{newell2016stacked} & CornerNet \cite{law2018cornernet} &Hourglass-104 \cite{newell2016stacked} &  40.5 &56.5& 43.1 &19.4& 42.7& 53.9& - & -\\
PANet\cite{liu2018path} & MaskRCNN \cite{he2017mask} & X101\cite{xie2017aggregated} & 46.6 & 65.1 & 50.6 & 29.3& 50.5 & 60.1 & 627 & 135\\
\hline
\hline
FPN \cite{lin2017feature} &  \multirow{2}{*}{FRCNN\cite{ren2015faster}} & \multirow{2}{*}{X101\cite{xie2017aggregated} }& 41.3 & 	62.5	 & 45.1	 & 24.2	 & 44.7 & 	51.7 &287 & 60 \\ 
TCC-FPN (ours) & &  & 43.0 &	64.0 &	47.0 &	24.2 &	45.9 & 55.3 & 246 & 59 \\ %
\hline
NasFPN\cite{ghiasi2019fpn} &\multirow{2}{*}{RetinaNet\cite{lin2017focal} }  & \multirow{2}{*}{R50$^*$\cite{he2016deep}} & 44.6& 	62.6& 	47.8& 	26.2& 	47.9& 	57.8& 555 & 60\\
TCC-NasFPN (ours)  & &  & 46.2& 	65.0& 	49.7& 	29.0& 	49.7& 	57.6 & 510 & 56\\
\hline
FPG\cite{chen2020feature}  & \multirow{2}{*}{FRCNN\cite{ren2015faster} } & \multirow{2}{*}{X101\cite{xie2017aggregated} } & 44.9&	64.3&	48.3&	24.6&	48.7&	58.9 & 686 & 100\\
TCC-FPG (ours) &  &  & 46.3&	66.9&	50.8&	28.7&	49.4&	56.5 & 626 & 100\\
\hline
RFP\cite{qiao2021detectors}  & \multirow{2}{*}{HTC\cite{chen2019hybrid} }  & \multirow{2}{*}{ViTAEv2-S\cite{zhang2022vitaev2}} & 53.1 & 72.2 & 57.7 & 32.8 & 56.2 & 68.3 & 548 & 115\\
TCC-RFP (ours)  &  &  & \textbf{53.5} & \textbf{72.6} & \textbf{58.2} & \textbf{33.3} & \textbf{56.7} & \textbf{68.6} & 455 & 114  \\
\hline
\multicolumn{3}{l}{\textsuperscript{*: Input images are resized to 1280 by 1280.}}
\end{tabular}
}
\label{tab:exp-all}
\end{table*}
\subsection{Setup}
For evaluation, we mainly use the popular MS COCO \cite{lin2014microsoft} dataset to help reveal the detection performance of different models. It has 118k training images, 5k validation images, and around 21k \textit{test-dev} images. We follow the MS COCO protocol and report the performance using the evaluation metrics of average precision (AP), AP at 0.5, AP at 0.75, and AP for small, medium, and large objects. The validation set is mainly used for ablation study, and the \textit{test-dev} set is used for main comparison. To present the analysis on computational complexities, we use the GFLOPs\footnote{GFLOPs: Giga floating point operations} as the main indicator. 

To validate the effectiveness of the proposed TCC, we apply the TCC on MFF for different popular and representative feature pyramids, such as FPN\cite{lin2017feature}, NasFPN\cite{ghiasi2019fpn}, FPG\cite{chen2020feature}, and RFP\cite{qiao2021detectors}. The details about these feature pyramids are listed as follows:
\begin{itemize}
    \item \textit{FPN}\cite{lin2017feature} is the initial and the most typical feature pyramid used in object detection. Its MFF is implemented after the fusion of higher-level feature and lower-level feature followed by a 3 by 3 convolution as refinement. To apply TCC in FPN, we simply replace the 3 by 3 convolution with the proposed TCC module. 
    \item \textit{NasFPN}\cite{ghiasi2019fpn} is derived from the FPN and introduces neural architecture search technology to obtain a novel feature pyramid structure. The MFF in the searched NasFPN structure is also based on operations like concatenation, sum, and convolution. The detailed implementation is discovered by the neural network by itself. The NasFPN is repeated several times to further refine the MFF results. In NasFPN, we apply the TCC after final MFF of each level and reduce one repetition of processing to save costs.
    \item \textit{FPG}\cite{chen2020feature} refers to the feature pyramid grid which introduces a deep multi-pathway feature pyramid by formulating the features from different levels as a regular grid of parallel bottom-up pathways. The MFF is performed by applying multi-directional lateral connections. It is not based on searching, but its structure is also repeated several times for refinement. Similar with NasFPN, we apply the TCC after final MFF of each level and also reduce one repetition.
    \item \textit{RFP}\cite{qiao2021detectors} means recursive feature pyramid, which incorporates additional feedback connections from FPN into the bottom-up backbone layers. It cooperates with a switchable atrous convolution for MFF. It is also similar with NasFPN ans FPG and applies multiple RFP modules to achieve advanced performance. In RFP, the original FPN structure is also applied. We thus can directly apply the TCC in FPN. IN addition, we also apply the TCC after the switchable atrous convolution for MFF.
\end{itemize}

It is worth mentioning that these feature pyramids are originally implemented in different detection pipelines such as Faster RCNN\cite{ren2015faster}, RetinaNet\cite{lin2017focal}, and HTC\cite{chen2019hybrid}. For fair comparison, we use the same detection pipeline when applying the proposed TCC in the corresponding feature pyramid. For training different feature pyramids, we follow the protocols reported in the related papers and reproduce all the baseline results using the publicly released code \footnote{https://github.com/open-mmlab/mmdetection}. 

\begin{table*}[t!]
\centering
\caption{Ablation study of different multi-level feature fusion methods in FPN on MS COCO \textit{val} set.}
\resizebox{\linewidth}{!}{
\begin{tabular}{l | c| c ccc  c| c |c }
\hline
MFF Refinement& AP & AP$_{50}$ & AP$_{75}$ & AP$_{S}$ & AP$_{M}$ & AP$_{L}$  & GFLOPs & Params(M) \\ 
\hline
No Refinement &36.9 &58.1 &40.0 &21.5 &40.7 &47.6& 156&40\\ 
Refine with Convolution 3 x 3  (Oiriginal FPN)& 37.4	&58.1&	40.4&	21.2&	41&	48.1 &207&42\\
\hline
Refine with Local Contexts Only &38.9	&60.1	&42.3	&22.9	& 42.6&	51.2&159&40\\
Refine with Condensed Contexts without Transformer & 39.3	&60.2&	42.9	&23	&24.7&	51.1 &161&41
\\
Refine with TCC&40.3& 61.8 & 43.5 & 23.7 & 43.9 & 52.6 &162&41\\ %
\hline
\end{tabular}
}
\label{tab:abl-1}
\end{table*}

% abl - 2 

\begin{table*}[t!]
\centering
\caption{ Effects of predicting and using different numbers of key global context features when condensing contexts in TCC.}
\resizebox{0.7\linewidth}{!}{
\begin{tabular}{c |  c| c ccc  c  } % | c |c
\hline
Number of Key Global Contexts  &AP & AP$_{50}$ & AP$_{75}$ & AP$_{S}$ & AP$_{M}$ & AP$_{L}$  \\ % & GFLOPs & Params(M) 
\hline
1 &39.8	&60.9&	43.2&	22.9&	43.4&	51.7\\ 
\hline
2 &40.2	&61.3&	43.9&	23.7&	43.5&	52.5\\
%\hline
4 &40.3& 61.8 & 43.5 & 23.7 & 43.9 & 52.6 \\ %
%\hline
8 &40.2 & 61.7 & 43.5 & 24.5 & 43.9 & 52.8 \\ 
\hline
\end{tabular}
}
\label{tab:abl-2}
\end{table*}

\begin{figure*}[t]
\includegraphics[width=\linewidth, height=0.45\textheight]{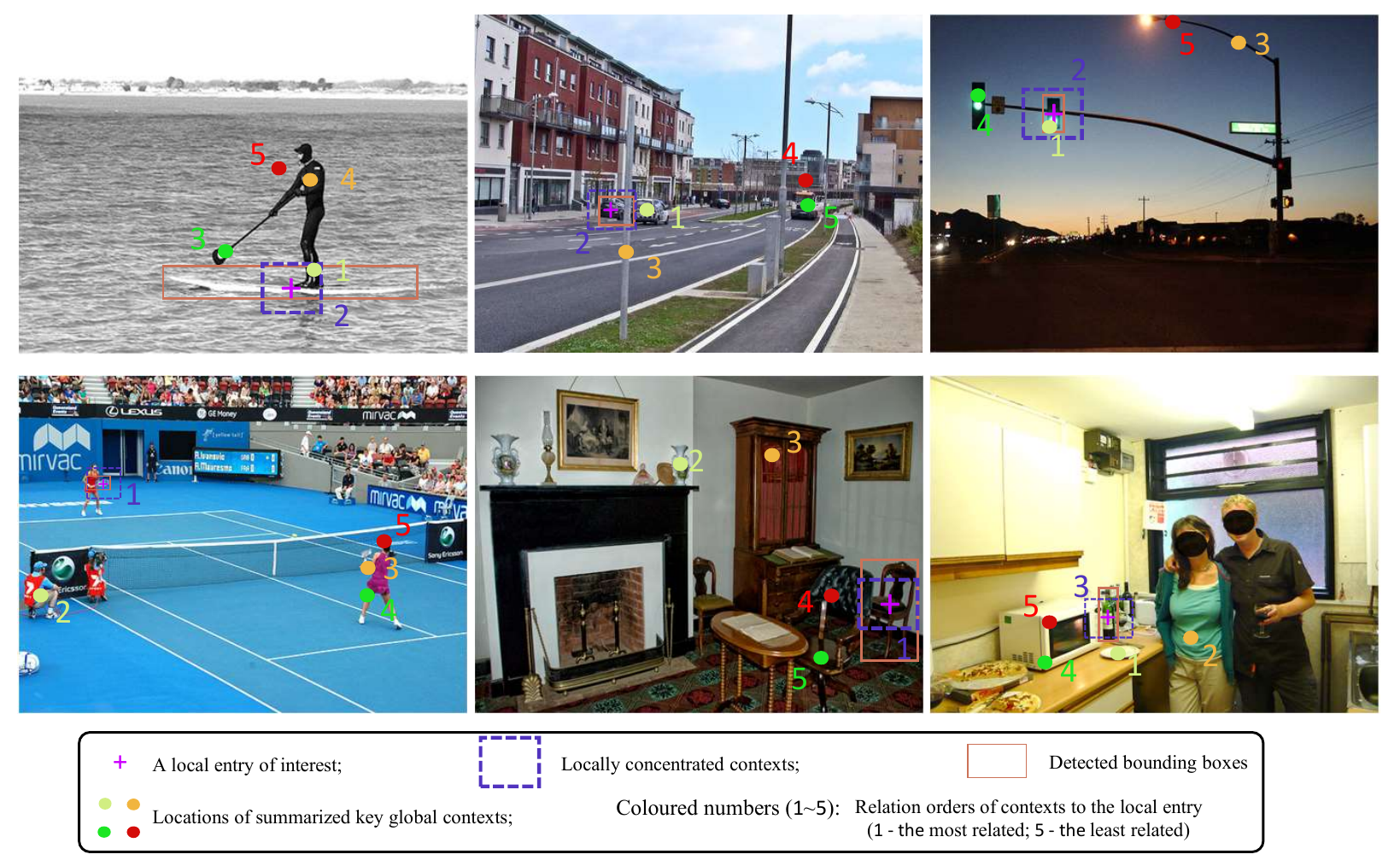}
  \caption{Visualization of the spatial arrangements of condensed contexts and their relations to an entity from a specific local entry. }%
  %\vspace{-0.3cm}
  \label{fig:kps}
\end{figure*}

\begin{figure*}[t]
\includegraphics[width=\linewidth,height=0.5\textheight]{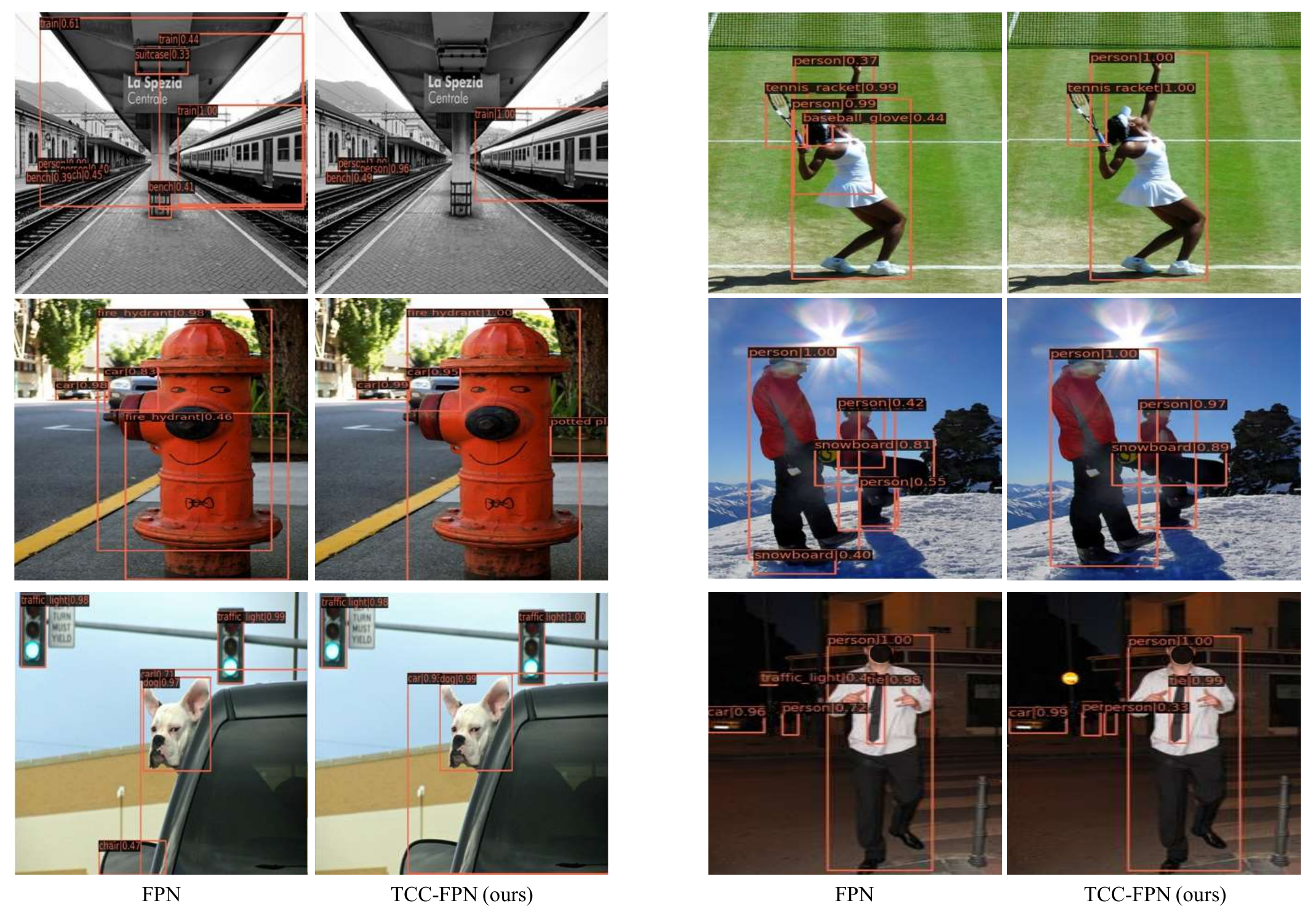}
  \caption{Visual examples of the detection results of TCC compared to the baseline FPN method. }%
  %\vspace{-0.3cm}
  \label{fig:qua}
\end{figure*}

\subsection{Detection Performance}
\subsubsection{Performance on Different Feature Pyramids}
Firstly, we study the improvement brought by TCC on different feature pyramids regarding the efficiency and accuracy. We also use different backbone networks to prove that our method is not sensitive to different implementation choices. Using the \textit{test-dev} set of MS COCO for evaluation. Fig. \ref{fig:ap-gflops} shows the AP results against GFLOPs.

From the presented results, we can observe that our TCC is able consistently improve the accuracy for different baseline feature pyramids and different backbone networks using less computational complexities. In particular, for a typical FPN\cite{lin2017feature}, the baseline method with ResNet50\cite{he2016deep} as backbone achieves an AP of 37.3 at a cost of 208 GFLOPs, while using our proposed TCC can help the FPN with ResNet50 achieve 40.6 at a cost of 163 GFLOPs, which means that the TCC improves AP with 3.3 points with 45 less GFLOPs. With a deeper backbone, ResNet101, the TCC also improves AP and reduces GFLOPs promisingly. 
For the NasFPN\cite{chen2020feature}, the baseline uses different input image sizes, \textit{e.g.} 640 by 640 or 1024 by 1024. Using TCC, we improve the AP of NasFPN from 39.9 and 44.1 to 41.6 and 45.4 for 640 and 1024 input image sizes, respectively. Correspondingly, the GFLOPs are reduced from 139 and 355 to 128 and 326, respectively. For the FPG\cite{chen2020feature}, there are similar improvements of using TCC, \textit{i.e.} increasing AP from 42.4 and 43.8 to 44.1 and 45.2, and decreasing GFLOPs from 602 and 681 to 539 and 518 for using ResNet50 and ResNet101 as backbone networks, respectively. For the RFP\cite{qiao2021detectors} that uses HTC\cite{chen2019hybrid} as detection pipeline, we can observe that our method still helps deliver improved performance with decreased GFLOPs and inference speed. For example, our method improves from 50.7 in AP to 51.5 in AP with a GFLOPs from 449 to 365. 

In addition to the GFLOPs, we also present the AP results against inference speeds of different methods in Fig. \ref{fig:speed}. All the speeds are obtained using the same hardware environment. Similar to the performance of AP against GFLOPs, we can observe that our method improves the AP and reduces the processing time, demonstrating the effectiveness of the proposed TCC for MFF in different feature pyramids.

\subsubsection{Results with High-performance Detectors}
In addition to the ResNet50 and ResNet101 based object detectors, we further introduce high-performance implementations to study the effects of TCC comparing to different state-of-the-art object detection algorithms like Cascade Mask RCNN\cite{he2017mask}, HTC\cite{chen2019hybrid}, FCOS\cite{tian2019fcos}, and so on. More specifically, for FPN and FPG that use Faster RCNN detection pipeline, we use the ResNeXt101\cite{xie2017aggregated} as the backbone network to improve detection accuracy. For the NasFPN that use RetinaNet, we use input image sizes of 1280 by 1280 to achieve high-performance detection. Then, for HTC-based RFP, we apply the most recently introduced ViTAEv2\cite{zhang2022vitaev2} as backbone network for comparison.
Table \ref{tab:exp-all} shows the results.  

According to the presented results, we can find that our method continues improving AP and reducing required GFLOPs at the same time, even on the much better implementations of detectors that may reach saturated performance. For example, the TCC improves FPN by 1.7 AP points and reduces around 40 GFLOPs. Similar improvements can also be observed for NasFPN and FPG. When using HTC-based RFP, we achieve 0.5 points' increase in AP and around 90 GFLOPs decrease in complexity comparing to the HTC-based RFP using ViTAEv2-S\cite{zhang2022vitaev2} as backbone. We would like to mention that the ViTAEv2 is a strong backbone network that can make the detector easily saturate for MS COCO. Nevertheless, we still improve the AP with TCC and our TCC can out-perform other compared state-of-the-art detectors when using the HTC-based RFP. This proves that our method is effective for improving the detection performance with contexts.

\subsection{Ablation study}
In this section, we study the effects of different components in the TCC module. We also present ablation studies on the numbers of predicted key global context locations. 

Table \ref{tab:abl-1} shows the performance of different MFF refinement methods and different components in the TCC. FPN with ResNet50 is used as the baseline, and \textit{val} set of MS COCO is used for evaluation. According to the results, we can find that the feature pyramid achieves degraded performance without refining MFF results. By subsequently adding the local contexts and summarized global contexts for the MFF in the feature pyramid, the detection performance is progressively improved. This illustrates that the contextual information is helpful for improving MFF in the feature pyramid and that the proposed condensed contexts are effective for collecting rich contextual information. Then, by further including the Transformer, our proposed TCC module achieves the highest accuracy and introduces smaller computational complexity than the baseline method.

In Table \ref{tab:abl-2}, we present the study of effects for using different numbers of summarized key global context features. We have studied the numbers ranging from 1 to 8, and the related detection accuracy and computational complexities are listed for comparison. According to the presented results, we can find that using 4 key global context features achieves the best speed/accuracy trade-off. Using 1 or 2 key global context features are beneficial but can be improved, while using 8 key global context features makes the detector saturates in accuracy. These results suggest that the global contexts are helpful for improving MFF in feature pyramids and they can be condensed into features from only a few key locations. It is worth mentioning that varying the number of considered key global contexts does not change the computational complexity too much, \textit{e.g.} using 8 key global context features would introduce less than 1 GFLOPs and less than 1 million (M) extra parameters.

\subsection{Qualitative Analysis} 
In addition to quantitative analysis, we also attempt to visualize the forms of contextual information condensed by the proposed TCC and compare our method with the baseline method visually. The FPN with ResNet50 is selected as the baseline in this section. 

Firstly, we present the spatial arrangements of learned locally concentrated and globally summarised contexts \textit{w.r.t.} the feature from a specific local entry. Fig. \ref{fig:kps} shows the visualization results. Note that the results of the last TCC on the deepest-level feature is selected for visualization. In this figure, we mainly present the locations of the specific local feature (pink cross), its locally concentrated contexts (purple dotted rectangular), and the summarized key global contexts (coloured dots), respectively. In addition to the locations, we also indicate the relations of different condensed contexts \textit{w.r.t.} the specific local feature. The relation orders (illustrated as coloured numbers: 1 for the most related; 5 for the least related) are also presented along side the corresponding contexts. From the figure, we can observe that the summarized key global contexts are generally located at meaningful instances that can represent the theme of the scene presented in the image. For example, in the top middle figure, the learned key global contexts are distributed on vehicles, buildings, and poles, which are reasonable to describe the urban scene of the image. Besides, the locations of summarized key global contexts, the relations between condensed contexts and the local feature learned by the Transformer in TCC also look natural. As an example, in the top-left figure, it shows that the TCC searches for human feet, paddles, and local surrounding visual information as the more useful cues for accurate detection of the surfboard, while the sea and the upper-body of the human are less important, which does not violate the common sense. These results have clearly demonstrated that the TCC can effectively model the relations between condensed contexts and local features and then help achieve effective detection based on the contexts and the modeled relations.

Besides the contexts, we further present the comparison of detection results in Fig. \ref{fig:qua}. According to the presented examples, we can find that our proposed TCC can help the baseline FPN improve the recognition quality promisingly. More specifically, on the top-left figure, the FPN with TCC can produce more accurate recognition of the train, avoiding many false positive detection results from the baseline FPN. In addition, we can observe the 2-nd row left figure and the top right figure, which show that our method helps the FPN provide more accurate detected bounding boxes to cover the objects more compactly. In other presented examples, we can find that the TCC continues to improve the FPN by enhancing the recognition quality and reducing the false alarms. These results demonstrate that our method is effective to help the detector understand the scene more robustly by introducing contexts.

\section{Conclusions}
In this study, we propose a Transformer-based context condensation module to achieve improved MFF results in different feature pyramids with a reduced complexity. By decomposing rich context information into a locally concentrated representation and a globally summarised representation, we obtain condensed contexts that can avoid exhausted context modeling to achieve promising refinement of fusion results. With the help of a Transformer decoder, our proposed TCC module can help boost detection accuracy with a smaller computational cost. In practice, we inject the TCC in to 4 different feature pyramids, including FPN, NasFPN, FPG, and RFP, we can observe consistent improvements, which validate the effectiveness of our method. We hope that our introduction of condensed contexts can open up a novel view of using contexts to improve visual understanding performance. 

\section{Data Availability}
The datasets generated during and/or analysed during the current study are available in the MS COCO repository, \url{https://cocodataset.org/}.

% \backmatter

% \section*{Declarations}

% Some journals require declarations to be submitted in a standardised format. Please check the Instructions for Authors of the journal to which you are submitting to see if you need to complete this section. If yes, your manuscript must contain the following sections under the heading `Declarations':

% \begin{itemize}
% \item Funding
% \item Conflict of interest/Competing interests (check journal-specific guidelines for which heading to use)
% \item Ethics approval 
% \item Consent to participate
% \item Consent for publication
% \item Availability of data and materials
% \item Code availability 
% \item Authors' contributions
% \end{itemize}

% \noindent
% If any of the sections are not relevant to your manuscript, please include the heading and write `Not applicable' for that section. 

%%===========================================================================================%%
%% If you are submitting to one of the Nature Portfolio journals, using the eJP submission   %%
%% system, please include the references within the manuscript file itself. You may do this  %%
%% by copying the reference list from your .bbl file, paste it into the main manuscript .tex %%
%% file, and delete the associated \verb+\bibliography+ commands.                            %%
%%===========================================================================================%%
\bibliographystyle{sn-basic} 
\bibliography{sn-bibliography}% common bib file
%% if required, the content of .bbl file can be included here once bbl is generated
%%\input sn-article.bbl

%% Default %%
%%\input sn-sample-bib.tex%

\end{document}